\documentclass{article}
\usepackage{spconf,amsmath,epsfig}
\usepackage{tikz}
\usepackage{amssymb}
\usepackage{tabularx}
\usepackage{pifont}
\usepackage{graphicx}
\usepackage{amsfonts}
\usepackage{booktabs}
\usepackage{makecell}
\usepackage{array}
\usepackage[square,numbers]{natbib}
\usepackage{caption, setspace}
\usepackage{soul}
\usepackage[activate={true,nocompatibility},final,tracking=true,kerning=true,spacing=true,factor=1100,stretch=10,shrink=10]{microtype}
\usepackage[hidelinks]{hyperref}

\title{A COMPARATIVE STUDY OF DEEP LEARNING LOSS FUNCTIONS\\FOR MULTI-LABEL REMOTE SENSING IMAGE CLASSIFICATION}
\name{Hichame Yessou, Gencer Sumbul, Beg\"{u}m Demir}
\address{Faculty of Electrical Engineering and Computer Science, Technische Universit\"at Berlin, Germany}

\begin{document}
\maketitle
\begin{abstract}
This paper analyzes and compares different deep learning loss functions in the framework of multi-label remote sensing (RS) image scene classification problems. We consider seven loss functions: 1) cross-entropy loss; 2) focal loss; 3) weighted cross-entropy loss; 4) Hamming loss; 5) Huber loss; 6) ranking loss; and 7) sparseMax loss. 
All the considered loss functions are analyzed for the first time in RS. After a theoretical analysis, an experimental analysis is carried out to compare the considered loss functions in terms of their: 1) overall accuracy; 2) class imbalance awareness (for which the number of samples associated to each class significantly varies); 3) convexibility and differentiability; and 4) learning efficiency (i.e., convergence speed). On the basis of our analysis, some guidelines are derived for a proper selection of a loss function in multi-label RS scene classification problems.
\end{abstract}

\vspace{-0.07in}
\section{Introduction}
\vspace{-0.07in}
\label{sec:intro}
Recent advances on remote sensing (RS) instruments have led to a significant growth of remote sensing (RS) image archives. Accordingly, multi-label image scene classification (MLC) that aims at automatically assigning multiple class labels (i.e., multi-labels) to each RS image scene in an archive has attracted great attention in RS. In recent years, deep learning (DL) based methods have been introduced for the MLC problems due to high generalization capabilities of DL models (e.g., convolutional neural networks (CNNs) and recurrent neural networks (RNNs)).
As an example, in~\cite{SigmoidCNN} conventional use of CNNs developed for single-label image classification is adapted for MLC. In this method, the sigmoid function is suggested for MLC adaptation instead of the softmax function as the activation of the last CNN layer. In~\cite{Stivaktakis:2019}, a data augmentation strategy is proposed to employ a shallow CNN in the framework of MLC. This method aims to apply an end-to-end training of the shallow CNN, while avoiding to use a pre-trained network. In~\cite{RsimCNN}, a multi-attention driven approach is introduced for high-dimensional high-spatial resolution RS images. In this approach, a branch-wise CNN is jointly exploited with an RNN to characterize a global image descriptor based on the extraction and exploitation of importance scores of image local areas. All the existing approaches utilize the conventional combination of sigmoid activation and cross-entropy loss functions to simultaneously learn multi-labels for each image in the framework of DL. Sigmoid activation function provides Bernoulli distributions and thus allows multiple class predictions. The cross-entropy loss function has strong foundations from information theory and its effectiveness has been widely proven. However, it is not fully suitable to use when: i) imbalanced training sets are present; and ii) there is a time constraint on the training phase of a DL based method. Since a loss function guides the whole learning procedure throughout the training, its proper selection is important for DL based MLC. 
Thus, in this paper, we present a study to analyze and compare different loss functions in the content of MLC and propose a scheme to guide the choice of loss functions based on a set of properties. All the considered loss functions are analyzed for the first time in RS in terms of their: 1) overall accuracy; 2) class imbalance awareness; 3) convexibility and differentiability; and 4) learning efficiency. BigEarthNet~\cite{BigEarthNet}, which is a large scale multi-label benchmark archive, is employed to validate our theoretical findings within experiments.
\vspace{-0.07in}
\section{Deep Learning loss Functions for Multi-Label Image Classification}
\vspace{-0.07in}
\label{sec:comparison}
Let $X = \{ x_1,...,x_M \}$ be an archive that consists of $M$ images, where $x_i$ is the $i$\textsuperscript{th} image in the archive. Each image in the archive is associated with one or more classes from a label set $\{l_1,...,l_C\}$. Let $y_{i,c}$ be a binary variable that indicates the presence or absence of a label $l_c$ for the image $x_i$. Thus, the multi-labels of the image are given by the binary vector $y_i=[y_{i,1},\ldots,y_{i,C}]$. A MLC task can be formulated as a function $F(x_i)= g(f(x_i))$ that maps the image $x_i$ to multiple classes based on the function $f(x_i) = p_i$ (which provides a classification score for each class in the label set) and the function $g(\cdot)$ (which defines the multi-labels of the image based on the probabilities). The learning process is performed by minimizing the empirical loss $L(y, y^*) = h(g(f(x_i)), y_i)$, which compares multi-label predictions with the ground reference samples. For a comparative analysis, we consider seven DL loss functions: cross-entropy loss (CEL)~\cite{Hinton1158}; focal loss (FL)~\cite{Focalloss}; weighted cross-entropy loss (W-CEL)~\cite{Hinton1158}; Hamming loss (HAL)~\cite{hammingloss}; Huber loss (HL)~\cite{huber1964}; ranking loss (RL)~\cite{rankingloss}; and sparseMax loss (SML)~\cite{sparseMax}. 
For the image $x_i$ we define its class probabilities $p_i$ as follows:
\begin{equation}
\label{p}
\setlength\abovedisplayskip{7pt}
\setlength\belowdisplayskip{7pt}
 p_i = \begin{cases}
 \hat{y}& \textit{if y = 1}\\
 1-\hat{y}&\textit{otherwise}
 \end{cases} 
\end{equation}
where $\hat{y}$ is resulting output from the Sigmoid activation function defined as $\delta(x) = 1/(1+e^{-x})$.
The CEL is formulated as:
\begin{equation}
\label{CE}
\setlength\abovedisplayskip{7pt}
\setlength\belowdisplayskip{7pt}
 \text{CEL} = -\sum log(p_i).
\end{equation}
For the CEL, easily classified images may significantly affect the value of the loss function and thus control the gradient that limits the learning from hard images. The FL adds a modulating factor to the CEL, shifting the objective from easy negatives to hard negatives by down-weighting the easily classified images as follows:
\begin{equation}
 \label{FL}
 \setlength\abovedisplayskip{7pt}
 \setlength\belowdisplayskip{7pt}
 \text{FL} = -\sum (1-p_i)^\gamma log(p_i) 
\end{equation}
where $\gamma$ is a focusing parameter, which increases the importance of correcting wrongly classified examples. Another way to guide the learning procedure is to consider class weighting that allows exploiting the importance for each class. The W-CEL is defined by setting a weighting vector inversely proportional to the class distribution. The HAL aims at reducing the fraction of the wrongly predicted labels compared to the total number of labels as follows:
\begin{equation}
 \label{HAL}
 \setlength\abovedisplayskip{7pt}
 \setlength\belowdisplayskip{7pt}
 \text{HAL}=\frac{1}{C}\sum_{c=1}^{C}y_{i,c} \oplus g(p_{i,c})
\end{equation}
where $\oplus$ denotes the XOR logical operation. The HL consists of: i) a quadratic function for values in the target proximity; and ii) a linear function for larger values as follows:
\begin{equation}
 \label{HL}
 \setlength\abovedisplayskip{7pt}
 \setlength\belowdisplayskip{7pt}
 \text{HL}=\sum_{c=1}^{C}\begin{cases}
 max(0,1-y_{i,c}z_{i,c})^2, & \!\!\text{for} \, \, y_{i,c} \, z_{i,c} \geq -1 \\
 -4 \, y_{i,c} \, z_{i,c}, & \!\!\text{otherwise}
 \end{cases}
\end{equation}
where $z_{i,c}$ is the class score (i.e., logit) of the label $c$ without applying any activation function. It is worth noting that to utilize the HL, the value of $y_{i}$ is replaced by $y_{i} \in \{-1,+1\}^C$. The SML is coupled with the sparseMax activation function that provides sparse distributions, while holding a separation margin for classification. Its generalization for the multi-label classification is defined as follows:
\begin{equation}
 \label{SML}
 \setlength\abovedisplayskip{7pt}
 \setlength\belowdisplayskip{7pt}
 \text{SML} = -y_i^T z_i +\frac{1}{2} \sum_{j \in S} (z_{i,j}^2-\tau ^2(z_i))+\frac{1}{2} \|y_i\|^2
\end{equation}
where $\tau$ is a thresholding function to define which class scores will be further leveraged (denoted as $S$) and the remaining class scores will be truncated to zero (for a detailed explanation, see \cite{sparseMax}). The RL aims to provide an accurate order of class probabilities, and thus assign higher probabilities to ground reference classes compared to others. This is achieved with pairwise comparisons as follows:
\begin{equation}
 \label{RL}
 \setlength\abovedisplayskip{7pt}
 \setlength\belowdisplayskip{7pt}
 \text{RL} = \sum_{v\notin y_i}\sum_{u\in y_i} max(0, \alpha + z_{i,v}-z_{i,u}))
\end{equation}
where $u$ is the ground reference class labels associated with the image $x_i$ and $v$ is the remaining labels from the label set of the archive.

\vspace{-0.07in}
\section{A Comparative Analysis}
\vspace{-0.07in}
\label{sec:meth}
We analyze and compare the above-mentioned loss functions in the framework of MLC based on their: 1) class imbalance awareness; 2) convexibility and differentiability; and 3) learning efficiency. Our analysis of DL loss functions under these criteria aims at providing a guideline to select the most appropriate loss function for MLC applications. Most of the operational RS applications include a degree of \textit{class imbalance}, which is associated to the fact that classes are not equally represented in the archive. This is more evident in the case of MLC. When the number of images for a given class is not sufficient in the training set, characterization of this class can be more difficult compared to others. This may lead to misclassification of images. To overcome this limitation, the modulating factor defined in (\ref{FL}) significantly down-weights the effect of well-classified images on the value of the loss function (e.g., when $p_{i} \,\to\, 1$, the modulating factor shrinks towards 0). Since the FL focuses more on hard samples, minority classes can be better characterized.
In addition to FL, W-CEL considers images with minority classes more than the vastly represented classes in the training set. This is due to the fact that the weighting vector applied to the loss function is inversely proportional to the class distribution.
The optimization problems of DL methods are generally non-convex, while convex properties exist in the trajectory of gradient minimizers~\cite{GoodFellow}.
The convexity of a DL loss function is an important property for an effective training procedure and better generalization capability. In addition to the convexity, another factor that supports the optimization of a loss function is its differentiability. It is worth noting that the differentiability is not a sufficient condition for guaranteeing the convergence to a global minimum. However, it is a required condition for providing a non-zero gradient back to the DL model during backpropagation. There are several strategies that allow the training of non-differentiable loss functions. However, these strategies may undesirably change the aim of loss functions and introduce additional complexity.
\begin{figure*}[t]
\centering
 \includegraphics[width=\linewidth]{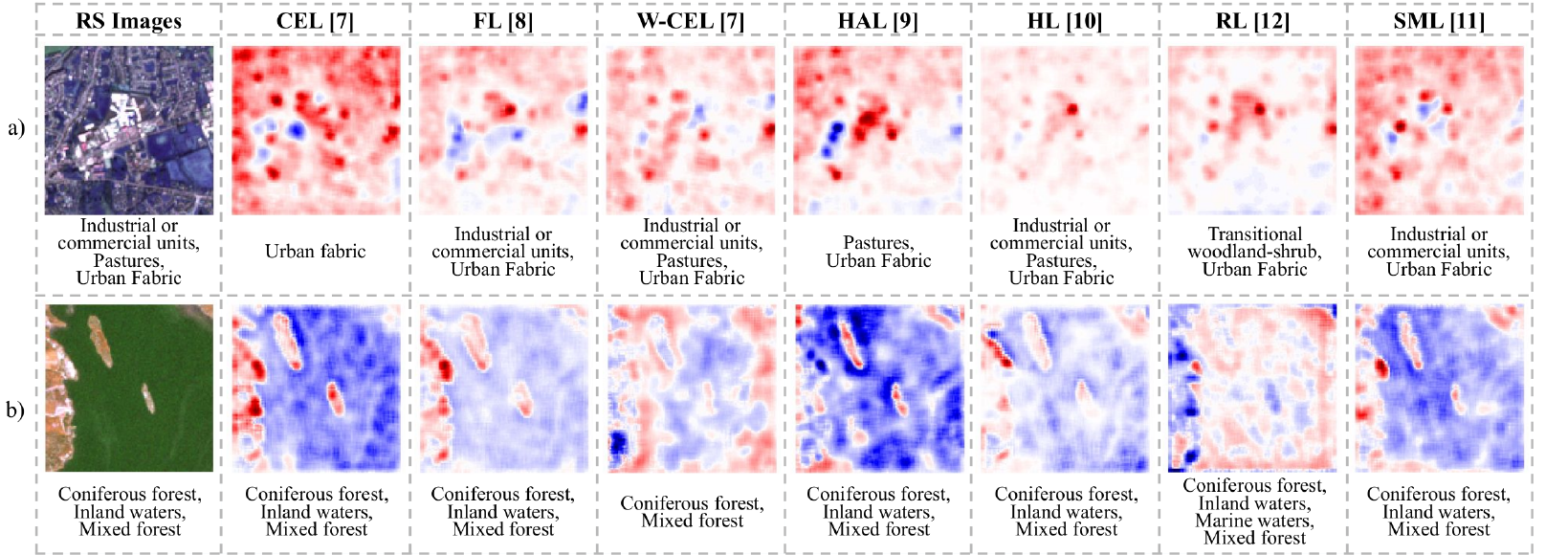}
 \caption{An example of the BigEarthNet images, their multi-labels and LRP heatmaps with the multi-label predictions of the considered loss functions. LRP heatmaps are given for the classes of a) Urban Fabric; and b) Coniferous Forest.}
 \label{fig:HMS}
 \vspace{-0.48cm}
\end{figure*}
Among the considered loss functions, only the HAL and RL do not embrace the \textit{convexity and differentiability}. This is due to the fact that they are non-convex and discontinuous, and thus difficult to be directly optimized.
The \textit{learning efficiency} is another criterion, which is evaluated as a rate at which the approximation of an iterative procedure in training reaches a high performance in terms of MLC. By employing more efficient learning procedures, similar MLC accuracies can be obtained with fewer iterations. Thus, a fast convergence reduces the total training time, which is required to reach a high MLC performance.
Accordingly, it is crucial for a DL loss function particularly when there is a time constraint on the training phase. In this work, we use the same optimization strategy for all loss functions, and thus do not assess the effect of optimizers on the learning efficiency.

\vspace{-0.07in}
\section{EXPERIMENTAL RESULTS}
\vspace{-0.07in}
\label{sec:typestyle}
\begin{figure}[t]
\centering
 \includegraphics[width=\linewidth]{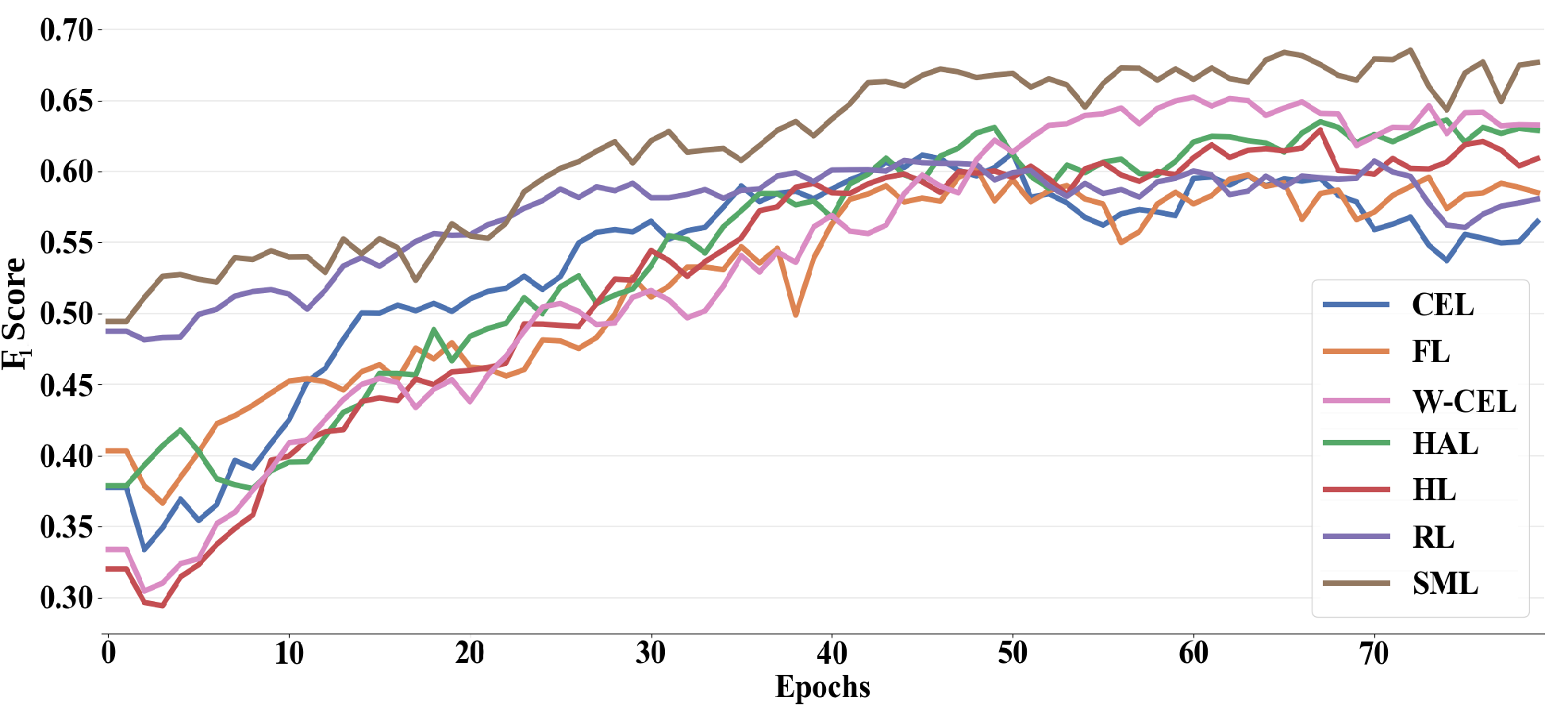}
 \caption{$F_1$-Scores over the validation set obtained by considering different loss functions during the different epochs of training.}
 \label{fig:F1Training}
 \vspace{-0.48cm}
\end{figure}
Experiments have been carried out on the BigEarthNet~\cite{BigEarthNet} large-scale benchmark archive. We used the BigEarthNet-19 class nomenclature proposed in~\cite{BigEarthNet19} instead of the original BigEarthNet classes. 
For the detailed explanation about the archive and the class nomenclature, the reader is referred to \cite{BigEarthNet} and \cite{BigEarthNet19}, respectively.
For the experiments, we considered a standard CNN architecture in order not to lose in generality. 
To this end, the CNN architecture given in the first step of the classification approach proposed in~\cite{RsimCNN} is used with the difference in terms of the number of units (1024) in the last two fully connected layers. We applied the same training procedure and hyperparameters to all considered loss functions for 80 epochs. Initial learning rate was selected as \(10^{-4}\) for the RMSprop optimizer. The performance of each loss function is provided in terms of precision ($P$), recall ($R$) and $F_1$-Score. We did not apply early-stopping with the validation set not to change the actual characteristics of the loss functions. We applied the Layer-wise Relevance Propagation (LRP)~\cite{LRP} technique to RGB spectral bands of the images. This technique allows propagating the multi-label predictions backward in CNNs and providing heatmaps, which indicate the most informative areas in RS images for each class. The heatmaps provide an accurate way to explain the characteristics of different loss functions. Low and high heatmap values are highlighted in blue and red tones, respectively.
\begin{table}[t]
\setlength{\tabcolsep}{4pt}
\renewcommand{\arraystretch}{0.1}
 \caption{Overall Precision, Recall and $F_1$-Score obtained using the CEL, FL, W-CEL, HAL, HL, RL and SML.}
\centering
\label{tab:scores}
\begin{tabular}{cccccccc}
\toprule
Metric & CEL & FL & W-CEL & HAL & HL & RL & SML \\ \midrule
$P$ ($\%$) & 75.2 & 72.2 & 76.2 & 75.9 & \textbf{76.6} & 58.0 & 70.7 \\ 
$R$ ($\%$) & 58.2 & 57.8 & 64.5 & 60.4 & 59.9 & \textbf{76.5} & 74.4 \\ 
$F_1$ ($\%$) & 62.3 & 61.1 & 66.2 & 64.2 & 64.1 & 62.9 & \textbf{69.9} \\ 
\bottomrule
\end{tabular}
\vspace{-0.48cm}
\end{table}
To analyze the \textit{overall accuracy} of the considered loss functions, Table \ref{tab:scores} shows the overall multi-label classification performances. As one can see from Table \ref{tab:scores}, the CNNs trained with HL and RL achieve the highest values of precision and recall, respectively. However, since the CNN trained with HL provides a low recall, it does not lead to a high $F_1$-Score. Similar to the HL, the CNN trained with RL leads to a low $F_1$-Score. Since the CNN trained with SML achieves high precision and recall, it leads to the highest $F_1$-Score compared to the other loss functions.
To analyze the \textit{class imbalance} and \textit{convexity and differentiability} criteria, Figure \ref{fig:HMS} shows two examples of the BigEarthNet images, their multi-labels and LRP heatmaps with multi-label predictions of the considered loss functions. From Fig. \ref{fig:HMS}.a, one can see the behavior of different loss functions when an image is associated with the classes, which are not equally represented in the archive. In detail, on the heatmap of the CEL, the semantic content associated with one of the well represented classes (which is Urban fabric) overwhelms the heatmap values. However, using the FL and W-CEL shows a more regular distribution of heatmap values. On the other hand, using the HAL and RL provides a high values associated with most of the image regions on the heatmap of the Urban fabric class while showing the highest values for the semantic content associated with the Industrial or commercial units class. In Fig. \ref{fig:HMS}.b, one can see that convex loss functions provide a more accurate distribution of heatmap values in terms of the correlation between the semantic content of the image and the heatmap values. Loss functions that hold convexity and differentiability have more reliable heatmap values. However, applying a weighting factor to a relatively smooth loss function such as the CEL introduces significant uncertainty in the heatmap values of W-CEL. In contrast to W-CEL, the modulating factor of the FL provides more regular values for the same regions. 
The RL and HAL show an irregular profile of predictions, while having high and low heatmap values associated with the same regions of the image. Although the LRP heatmaps are given for two examples, the similar behavior is also observed by varying the images in the BigEarthNet. To compare the \textit{learning efficiency} of the considered loss functions, Figure \ref{fig:F1Training} shows the overall $F_1$ scores on the validation set at different epochs of the training phase. As one can see from Figure \ref{fig:F1Training}, the CNNs trained with the SML and RL lead considerably better performances in $F_1$-Score from the initial epochs compared to the other loss functions. 
\begin{table}[!t]
\label{properties}
\small
\setlength{\tabcolsep}{3pt}
 \aboverulesep=0.5ex
 \belowrulesep=0.5ex
\renewcommand{\arraystretch}{1.}
 \caption{Comparison of the considered MLC loss functions. Different marks are provided: "H" (High), "M" (Medium), "L" (Low) or “NA” (Not Applied).}
\centering
\label{tab:properties}
\begin{tabular}{@{\hspace{-0.01cm}}l|c|c|c|c}
\toprule
\hline
\parbox[c]{1.35cm}{\centering Loss Function} & \parbox[c]{1.35cm}{\vspace{0.1cm}\centering Overall\\Accuracy\vspace{0.1cm}} & \parbox[c]{1.35cm}{\vspace{0.1cm}\centering Class\\ Imbalance\\Awareness\vspace{0.1cm}} & \parbox[c]{2cm}{\centering Convexity and\\Differentiability} & \parbox[c]{1.3cm}{\centering Learning Efficiency} \\
\hline
CEL~\cite{Hinton1158} & $\mathrm{L}$ & $\mathrm{L}$ & $\mathrm{M}$ & $\mathrm{M}$ \\ \hline
FL~\cite{Focalloss} & $\mathrm{L}$ & $\mathrm{H}$ & $\mathrm{M}$ & $\mathrm{L}$ \\ \hline
W-CEL~\cite{Hinton1158} & $\mathrm{M}$ & $\mathrm{H}$ & $\mathrm{M}$ & $\mathrm{L}$ \\ \hline
HAL~\cite{hammingloss} & $\mathrm{M}$ & $\mathrm{L}$ & $\mathrm{NA}$ & $\mathrm{M}$ \\ \hline
HL~\cite{huber1964} & $\mathrm{M}$ & $\mathrm{L}$ & $\mathrm{H}$ & $\mathrm{M}$ \\ \hline
RL~\cite{rankingloss} & $\mathrm{L}$ & $\mathrm{L}$ & $\mathrm{NA}$ & $\mathrm{H}$ \\ \hline
SML~\cite{sparseMax} & $\mathrm{H}$ & $\mathrm{M}$ & $\mathrm{M}$ & $\mathrm{H}$ \\ \hline
\bottomrule
\end{tabular}
\vspace{-0.42cm}
\end{table}
\vspace{-0.07in}
\section{CONCLUSION}
\vspace{-0.07in}
\label{sec:conc}
This paper analyzes and compares different loss functions in the framework of MLC problems in RS. In particular, we have presented advantages and limitations of different DL loss functions in terms of their: 1) overall accuracy; 2) class imbalance awareness; 3) convexity and differentiability; and 4) learning efficiency. In Table \ref{tab:properties}, a comparison of the considered loss functions is given on the basis of our experimental and theoretical analysis. In greater detail, experimental results show that the highest overall accuracy is achieved when the SML is utilized as a loss function. The FL and W-CEL can be more convenient to be utilized as loss functions when the imbalanced training sets are present. For the MLC applications that require a training phase with convex and differentiable loss functions, the HAL and the RL are less suitable to be used during the training phase. The SML and RL can be more convenient to be utilized as loss functions when a lower computational time is preferred for the training phase of a DL based MLC method. This study shows that for MLC problems in RS, DL loss functions should be chosen according to the need of the considered problem. As a future work, we plan to further analyze the differences of the MLC loss functions by visualizing their 
3D trajectories under different network architectures.
\vspace{-0.23in}
\section{\small Acknowledgements}
\vspace{-0.07in}
\small
This work is funded by the European Research Council (ERC) through the ERC-2017-STG BigEarth Project under Grant 759764.
\bibliographystyle{IEEEbib}
\bibliography{defs, refs}
\end{document}